\documentclass[letterpaper, 10 pt, conference]{ieeeconf}  
\IEEEoverridecommandlockouts                              
\usepackage[english]{babel}
\usepackage[T1]{fontenc}

\usepackage[nolist]{acronym}
\usepackage[cmex10]{amsmath}
\usepackage[font={small}]{caption}
\usepackage{subcaption}
\usepackage{amssymb}
\usepackage{bm}
\usepackage{booktabs}
\usepackage{color}
\usepackage{float}
\usepackage{graphicx}
\usepackage{hyperref}
\usepackage{import}
\usepackage{mathtools}
\usepackage{multirow}
\usepackage{multicol}
\usepackage{flushend}
\usepackage{outline}
\usepackage{paralist}
\usepackage{siunitx}
\usepackage{stmaryrd}
\usepackage{systeme}
\usepackage{units}
\usepackage{url}
\usepackage{tikz}
\usepackage{booktabs}
\usepackage{balance}
\usepackage{cite}
\usetikzlibrary{tikzmark}
\usepackage{esvect}
\usepackage[normalem]{ulem}

\title{\LARGE \bf Hierarchical Reinforcement Learning for Precise Soccer Shooting Skills using a Quadrupedal Robot}
\author{Yandong Ji$^{*,1}$, Zhongyu Li$^{*,1}$, Yinan Sun$^{1}$, Xue Bin Peng$^{1}$, Sergey Levine$^{1}$, Glen Berseth$^{2}$, Koushil Sreenath$^{1}$
\thanks{$^*$ Authors contributed equally.}
\thanks{$^1$ University of California, Berkeley. \{yandong0204, zhongyu\_li, syn1122, xbpeng, koushils\}@berkeley.edu, svlevine@eecs.berkeley.edu}
\thanks{$^2$ Universit\'e de Montr\'eal, Mila, glen.berseth@mila.quebec}
}

\begin{document}
\maketitle

\begin{abstract}
We address the problem of enabling quadrupedal robots to perform precise shooting skills in the real world using reinforcement learning. 
Developing algorithms to enable a legged robot to shoot a soccer ball to a given target is a challenging problem that combines robot motion control and planning into one task. 
To solve this problem, we need to consider the dynamics limitation and motion stability during the control of a dynamic legged robot. 
Moreover, we need to consider motion planning to shoot the hard-to-model deformable ball rolling on the ground with uncertain friction to a desired location.
In this paper, we propose a hierarchical framework that leverages deep reinforcement learning to train (a) a robust motion control policy that can track arbitrary motions and (b) a planning policy to decide the desired kicking motion to shoot a soccer ball to a target. 
We deploy the proposed framework on an A1 quadrupedal robot and enable it to accurately shoot the ball to random targets in the real world. 
\end{abstract}

\section{Introduction}
\label{sec:intro}

\begin{figure}[t]
    \centering
    \includegraphics[width=0.9\linewidth]{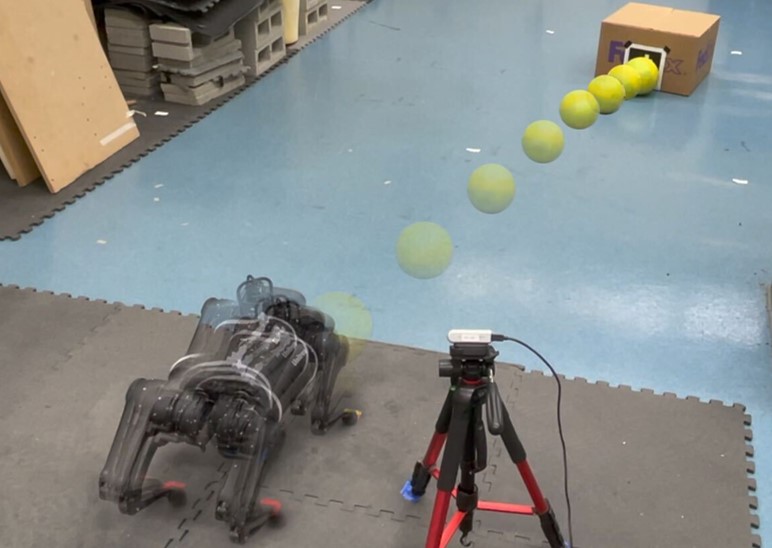}
    \caption{A quadrupedal robot precisely shoots a soccer ball to a randomly given target using the proposed hierarchical RL framework which includes a motion control policy and a motion planning policy. The results are best seen at the video: \url{https://youtu.be/bteipHcJ8BM}.}
    \label{fig:main}
    \vspace{-0.5cm}
\end{figure}

Quadrupedal robots have attracted a great deal of interest in the robotics community, and classical model-based methods~\cite{winkler2018gait,kim2019highly,gilroy2021autonomous} have been effective paradigms for designing controllers for these robots. 
However, model-based methods need careful system modeling, with intricate models being less practical for online deployment due to computational limits. 
Furthermore, it can be difficult to apply model-based methods in settings where the dynamics are difficult to model, such as a deformable soccer ball.
To shoot a ball, a controller is needed to enable the robot to swing its leg quickly to gain enough momentum to kick the ball to roll on the ground while also maintaining balance of its body during the fast kicking motion. 
Besides the motion control of the robot, motion planning is also challenging and requires the robot to consider the real-time ball and goal positions to generate a reasonable body motion while respecting the robot's physical limits.
Moreover, in order to precisely shoot the ball to the goal, the motion planner needs to also deal with the hard-to-model contact between the robot and the deformable soccer ball, as well as uncertainties associated with the rolling friction between the ball and the ground.
Although this task can be challenging for model-based methods, recent deep reinforcement learning~(RL) methods have presented potential techniques for tackling this without the need for explicit models of systems.
In this paper, we develop a hierarchical model-free RL framework that includes a motion control policy and a planning policy to perform precise shooting skills on quadrupedal robots in the real world, as shown in Fig.~\ref{fig:main}.

\subsection{Related Work}

There has been exciting recent progress on applying deep RL on nonprehensile manipulation tasks using robotic arms and locomotion tasks with legged robots, respectively. In this section, we review the most relevant work in these domains.

\subsubsection{Nonprehensile Manipulation using Robotic Arms}
Deep RL has been used to train robotic arms to hit, throw, or toss objects in both simulated and real-world environments.
Controllers have been trained to precisely hit a moving object in the air~\cite{PGPeters2008,PetersTenis2013,deisenroth2014multi}, as well as to throw an object to a target~\cite{DPPT2017}.  
For example, in~\cite{KoberDart2011}, motor primitives are learned by imitation learning and RL for a dart throwing task and for hitting flying table tennis balls. 
For tossing objects to a target, both model-based RL and model-free RL techniques have been utilized in~\cite{zeng2019tossingbot}.
However, most of these prior systems focus on tasks that primarily involve ballistic motion of the objects. 
For more contact-rich tasks, such as kicking a soccer ball rolling on the ground, the movement of the object involves complex rolling friction and contact dynamics that are substantially harder to control. 
There are recent endeavors using guided policy search to train robots to hit a hockey puck on the ground~\cite{ChebotarHockey2017} and using Bayesian system identification to play mini golf~\cite{muratore2022neural}.
However, the objects used in these work are primarily rigid, with simpler contact dynamics compared to deformable objects, such as a soft soccer ball. 
Furthermore, all of the aforementioned systems use fully actuated robotic arms with static bases.
The problem becomes more difficult when controlling a legged robot, since the controller now must also maintain stability of a robot with a floating base, which is not considered by previous work.
For example, a quadrupedal robot learns to manipulate a soft ball while laying the robot base on the ground to avoid considering the balancing problem in~\cite{shi2021circus}.

\subsubsection{Locomotion on Quadrupedal Robots with RL}
There has a been a large body of recent work on applying model-free RL to control quadrupedal robots for agile locomotion skills in the real world~\cite{hwangbo2019learning,peng2020learning,kumar2021rma,smith2021legged,ji2022concurrent}.
Many of these systems use some form of sim-to-real transfer, which allows policies trained in simulation to be deployed on a real robot.
These sim-to-real techniques can be broadly classified into two categories: zero-shot transfer by training a policy for direct deployment on the hardware~\cite{peng2018sim,hwangbo2019learning,li2021reinforcement,ji2022concurrent,kumar2021rma}, and methods that continue to incorporate real world data~\cite{peng2020learning,smith2021legged} after pretraining in simulation.
However, all of above-mentioned RL approaches focus on developing low-level locomotive skills, such as walking.
Using quadrupeds to accomplish a more complex and higher level tasks, such as combining the robot's body motion control and motion planning to shoot a deformable object, is not addressed. 

\subsubsection{Legged Soccer Robots}

Developing a legged robot that can kick a ball like a human has attracted lot of interest in the robotics community, and notably, in the RoboCup Leagues~\cite{stone2007intelligent}. 
Most previous approaches tackling this problem focus more on rule-based motion primitives, without considering shooting to a desired location, such as the work using quadrupedal robots~\cite{veloso1998playing, chalup2007machine, stone2007intelligent, cherubini2010policy} and humanoid robots~\cite{friedmann2008versatile, behnke2008hierarchical,acosta2008modular} in the RoboCup. 
There have been some attempts to improve shooting skills, such as~\cite{jouandeau2014optimization} where trajectory optimization is utilized to improve shooting distance, but this is only validated in simulation. 
Recently, model-free RL has demonstrated promising results for training a single policy to dribble, kick, and shoot a soccer ball with simulated humanoid robots~\cite{peng2017deeploco,teixeira2020humanoid,da2021deep}.
These end-to-end methods have the advantage of being able to produce more agile behaviors by directly leveraging the robot's full-order dynamics and contacts with the ball.
However, transferring policies learned in simulation to real legged robots is challenging due to differences between the dynamics of the simulator and the physical system.
Therefore, effectively leveraging RL to develop soccer shooting skills for legged robots in the real world remains an open problem and has yet to be demonstrated on a real-world legged robot.

\subsection{Contributions}
The central contribution of this work is the design and development of a hierarchical reinforcement learning frameworks for precise soccer shooting skills using quadrupedal robots.
This framework leverages model-free RL to enable a standing quadrupedal robot to precisely shoot a soccer ball by coupling robot motion control and motion planning.
Our motion control policy learns various full-body motions in order to track random parametric end-effector (toe) trajectories while maintaining balance during standing.
The motion planning policy is responsible for shooting the deformable soccer ball to a desired target. This uses a multi-stage training approach, wherein a planning policy is first trained with a rigid ball in simulation, then fine-tuned with a soft ball in the real world. 
We validate the proposed methodology in real-world experiments using a quadrupedal robot, and demonstrate the feasibility of attaining a robust control policy for dynamic soccer shooting motions and precise planning policy to shoot a deformable soccer ball to a random specified target in a relatively large range. 
This paper serves as a step towards the development of RL-based quadrupedal robotic soccer players in the real world that could one day compete with humans.

\label{sec:Introduction}

\section{Soccer Shooting Skill using a Quadruped}
\label{sec:prelim}

\begin{figure}[t]
    \centering
    \includegraphics[width=0.95\linewidth]{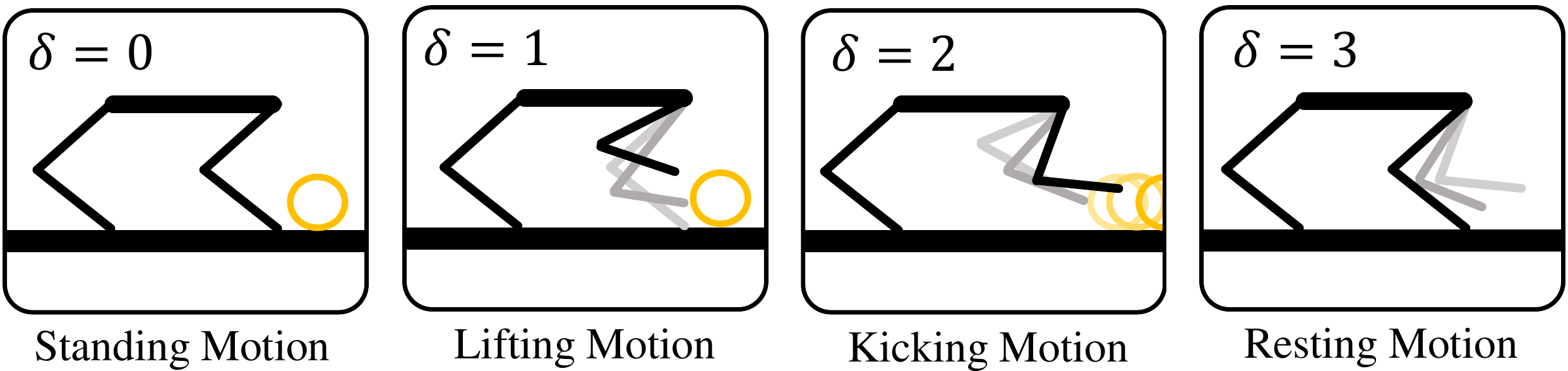}
    \caption{Motions that constitutes the shooting maneuver of a soccer ball (marked by the yellow circle). The quadrupedal robot needs to start with a standing motion, then lift its leg up and hold it in the air to prepare for the kicking motion once commanded. After kicking, the robot should put the leg down and get back to standing.}
    \label{fig:motion_breakdown}
    \vspace{-0.5cm}
\end{figure}

In this section, we provide a brief overview of the A1 quadrupedal robot, which serves as the experimental platform for this work. 
We then present our proposed approach for developing soccer shooting skills using the A1 robot.

\subsection{A1 Robot}
The A1 quadrupedal robot, shown in Fig.~\ref{fig:main}, has a total of 18 Degree-of-Freedoms~(DoFs). 
There are 6 DoFs for its base that include the saggital $q_x$, lateral $q_y$, vertical $q_z$ translational positions and roll $q_\psi$, pitch $q_\phi$, and yaw $q_\theta$ rotational positions. 
Each of the four legs has three actuated motors and we denote these joint coordinates as $q_m \in \mathbb{R}^{12}$.

\subsection{Soccer Shooting Skill}~\label{subsec:shooting_skill}
Our goal is to develop controllers that allow a quadrupedal robot to use one leg to shoot a ball to a specified target while standing. 
As illustrated in Fig.~\ref{fig:motion_breakdown}, we breakdown the entire shooting maneuver into four motions denoted by an indicator $\delta \in \mathbb{Z}$. 
The robot starts from a standing motion $\delta=0$ where all of the four legs are on the ground. 
Then, the robot needs to execute a lifting motion, denoted by $\delta=1$, where it lifts up one of its legs and holds it in position to prepare for the kick. 
We termed the leg that is lifted and used for kicking as the \textit{kicking leg}, while other legs are \textit{stance legs}. 
When the robot is commanded to kick, the robot executes a kicking motion, denoted by $\delta=2$, where the kicking leg should accelerate in order to gain momentum and then transfer the momentum to the ball on contact with the ball. 
The robot should also adjust the contact direction and contact force with the ball in order to shoot the ball to the specified target location. 
This kicking motion is difficult to perform on a quadrupedal robot, as the robot is not only required to transit from a stationary motion to a fast motion and slow down again in a very short time span, but also needs to prevent itself from falling over while executing fast kicks.
When the kicking is completed, the robot needs to put down the kicking leg and transit back to the nominal standing motion and get ready for the next round of kicks.
We refer to this last motion as the resting motion ($\delta=3$).
Each of these motions has its own time-span which is denoted as $T_\delta$.

\begin{figure*}[t]
    \centering
    \includegraphics[width=0.7\linewidth]{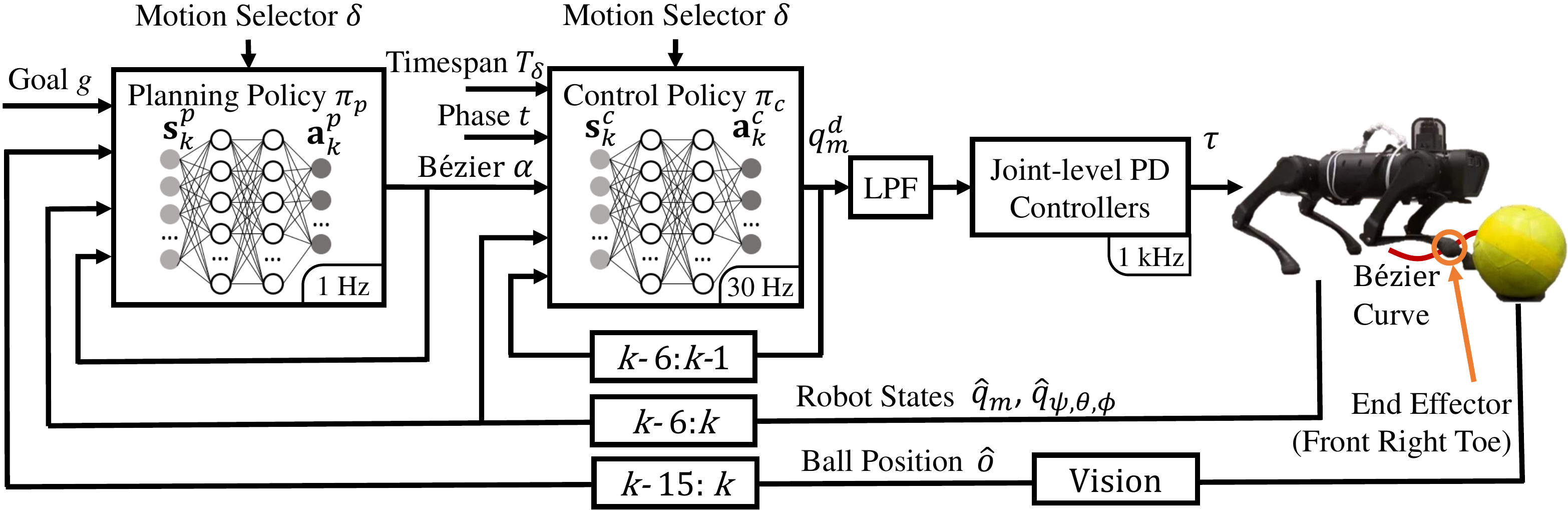}
    \caption{Proposed hierarchical reinforcement learning framework for a quadrupedal robot to perform a precise shooting maneuver. A control policy $\pi_c$ is firstly trained to track arbitrary end-effector (front right toe) trajectories represented by B\'ezier parameters $\alpha$ and motion time span $T_\delta$. The control policy replans at $30$ Hz. After obtaining a control policy that can reliably work on the real robot, we use it to train a planning policy $\pi_p$ to output a desired shooting motion to the controller in order to shoot the soccer ball to the goal $g$. The goal can be randomly placed and identified by an AprilTag~\cite{apriltag2011}. The planning policy updates its observation and action at $1$ Hz and is synchronized with the controller, \textit{e.g.}, the time step $k$ spans $1/30$ second. There is a rule-based motion selector to output an indicator $\delta$ to inform the planner and controller about the current motion type: standing, lifting, kicking, or resting.}
    \label{fig:framework}
    \vspace{-0.5cm}
\end{figure*}

\subsection{Motion Representation with B\'ezier Curves}
In this paper, we will use B\'ezier curves~\cite{choi2008path} to represent each motion ($\delta = 0,1,2,3$) of the soccer shooting maneuver for the control policy to perform, and will have a planning policy to compute the parameters of the desired B\'ezier curve. 
We do this to take into account various features of parametric B\'ezier curves, such as smoothness of the curve and physical meaning of the parameters. 

We use the Front Right~(FR) leg of the robot as the kicking leg. 
If we consider the robot's kicking leg as a robot arm attached to the robot's floating base, then the toe of the kicking leg can be considered as the \textit{end-effector}.
We define the toe position of the kicking leg as $x_e\in \mathbb{R}^3$. 
The trajectory of the end-effector in the $3D$ space can be parameterized by a B\'ezier curve~\cite{choi2008path},
\begin{equation}~\label{eq:bezier}
    B_\alpha(t) = \sum_{i=0}^{n} \frac{n!}{i!(n-i)!} (1-t)^{n-i}t^i \alpha_i
\end{equation}
where $\alpha_i \in \mathbb{R}^{3\times1}$ are the B\'ezier parameters, with $n+1$ being the number of the parameters. We use $n=4$ in this paper. 
The variable $t \in [0,1]$ is the phase time that is scaled by the timespan of the trajectory $T_\delta$.
Therefore, the end-effector trajectory can be defined by a discrete set $\alpha = \{\alpha_0, \alpha_1, \dots, \alpha_4\}$. 
Please note that, the resulting B\'ezeir curve is infinitely differentiable with respect to phase $t$ and we can not only obtain the reference position information from the curve but also higher order terms such as reference velocity and acceleration.   
Furthermore, the four motions discussed in Fig.~\ref{fig:motion_breakdown} can have different B\'ezier parameters $\alpha$.
Thus, enabling the robot to perform different motions in the shooting maneuver can now be generalized to tracking B\'ezier curves with different parameters.

Given the above end-effector trajectory for the swing leg with the constraint that the stance legs have their toes on the ground, one can typically use Inverse Kinematics (IK) to solve for all the leg joints and the body pose. 
However, as we will see in the following sections, instead of solving IK online, we can train a policy to directly move the robot so as to directly follow the reference swing leg end-effector trajectory specified by the B\'ezier parameters.

\section{Hierarchical Learning Framework}
\label{sec:framework}

We now propose a framework that leverages hierarchical reinforcement learning to enable a quadrupedal robot to learn soccer shooting skills.
As illustrated in Fig.~\ref{fig:framework}, we break down the soccer shooting problem into two parts: to plan for the robot's shooting motion in order to kick the ball to a given target, and to control the robot to perform the planned motion while staying balanced.

A control policy $\pi_c$ is first developed to enable the robot to track arbitrary end-effector (FR toe) trajectories while preventing falling by adjusting the robot pose via stance legs. 
This is realized by training a policy to output desired robot motor positions $q^d_m$ in order to track a desired end-effector trajectory represented by B\'ezier parameters $\alpha$ while respecting the robot dynamics. 
This control policy runs at $30$ Hz and its output $q^d_m$ is passed through a Low Pass Filter~(LPF)~\cite{peng2020learning,li2021reinforcement} and joint-level PD controllers to obtain desired motor torques $\tau$ on the robot at $1$ kHz. 

The soccer ball position $\hat{o} \in \mathbb{R}^3$ in the robot body frame is detected by a RGB-Depth camera, and in order to shoot the soccer to a given goal location $g \in \mathbb{R}^2$, we develop a planning policy $\pi_p$ on top of the control policy.
This planning policy is trained to examine the current ball position, robot states, and the shooting goal, and based on these, to generate optimal end-effector B\'ezier parameters for the controller to track.  
This planning policy replans at $1$ Hz to deal with online disturbance and tracking errors due to the controller.
The planning policy developed in this work only considers shooting the ball to a target rather than to enable the ball to follow a trajectory while reaching the target.

There is a rule-based motion selector to indicate to the robot what motion presented in Fig.~\ref{fig:motion_breakdown} should be performed at the current time. 
The appropriate motion indicator $\delta$ will be selected and passed to both the planning and control policies.  

The advantage of using this hierarchical learning framework is that we can separate the high-level shooting skills into two subproblems: control and planning. 
We can first focus on training a robust control policy in simulation that can be transferred from simulation to the real world to allow the robot to track arbitrary end-effector trajectories without causing the robot to fall over.
Afterwards, we can reuse this control policy for developing the planning policy and focus only on how to precisely shoot the soccer to the goal.

\section{Learning the Soccer Shooting Control}
\label{sec:control}
We now present the development of the control policy, which is first trained in simulation by RL and then directly transferred to a real robot, allowing the robot to track arbitrary end-effector trajectories while balancing. 

\subsection{Training Environment}
The environment for training the control policy for the A1 quadrupedal robot agent is developed in MuJoCo~\cite{todorov2012mujoco}. 
The details of the training environment are introduced below. 

\subsubsection{Action Space}
The action $\mathbf{a}^c_k$ of control policy at time step $k$ is the desired joint position $q^d_m \in \mathbb{R}^{12}$.
These are passed through a Low Pass Filter (LPF) and input to joint-level PD controllers to obtain the motor torques $\tau \in \mathbb{R}^{12}$, as shown in Fig.~\ref{fig:framework}.

\subsubsection{State Space}
As illustrated in Fig.~\ref{fig:framework}, the observation $\mathbf{s}^c_k$ of the control policy at time step $k$ contains 6 parts.
The first part is the current motion indicator $\delta \in \mathbb{Z}$ from the motion selector. 
As introduced in Sec.~\ref{subsec:shooting_skill}, $\delta$ is selected from $\{0,1,2,3\}$ that represents standing, lifting, kicking, and resting motions, respectively.
This indicator can help the robot to understand and distinguish the different motions that it is required to perform.  
The second and third parts are the desired B\'ezier parameters $\alpha \in \mathbb{R}^{3\times5}$ and corresponding trajectory timespan $T_\delta$, respectively.
The entire desired trajectory can then be well defined by these two variables.
The fourth input is the current motion phase $t \in [0,1]$ scaled by timespan $T_\delta$, which implicitly informs the robot about the desired end-effector position. 
The fifth part of the observation consists of the robot's current and past states.
The feedback of the robot states at time step $k$ includes measured robot joint position $\hat{q}_m \in \mathbb{R}^{12}$ and robot orientation $\hat{q}_{\psi,\theta,\phi} \in \mathbb{R}^3$. 
Please note that this feedback is raw sensor reading and we do not require a state estimator or other filters. 
Instead, we include a history of the past $6$ time steps (around $0.2$ seconds) of robot state feedback into the observation and encourage the policy to learn the filtering and state estimation by itself.
Finally, the sixth part of the observation is a history of the past $6$ time steps of previous action $\mathbf{a}^c_{k-1:k-6}$.
We hypothesize that the history of past robot states and actions enables the control policy to infer the closed-loop dynamics of the robot using the input/output data.

\subsubsection{Policy Representation}
The control policy is represented by a deep neural network that consists of two hidden layers. 
Each one is a fully-connected layer with $512$ units and $tanh$ activation. 

\subsection{Reward}
At each time step, the robot takes an observation, executes one action obtained by the policy, and receives a reward. 
The reward function for the control policy is designed to encourage the robot to follow the desired end-effector trajectory while maintaining balance and improving the smoothness of the robot motions.
The reward $r_{c,k} \in [0,1]$ at time step $k$ is formulated as:
\begin{equation}~\label{eq:control_reward}
    r_{c,k} = w_c^T [r^e_{c,k}, r^m_{c,k}, r^{\dot{m}}_{c,k}, r^b_{c,k}, r^{\dot{b}}_{c,k}, r^{\tau}_{c,k}, r^{\Delta \mathbf{a}}_{c,k}]^T.
\end{equation}
Here, $r^e_{c,k}$ stands for the reward in terms of the end-effector position tracking error and is formulated as
\begin{equation}~\label{eq:end_effector}
    r^e_{c,k} = \exp({-\rho^e||x_{e,k} - B_{\alpha}(t)||_2^2}),
\end{equation}
where $x_{e,k}$ is the current ground truth end-effector (front right toe) position at time step $k$, $B_{\alpha}(t)$ is the reference end-effector position calculated via~\eqref{eq:bezier}, and $\rho^e >0$ is a scaling variable. 
This term is designed to encourage the robot's front right toe (end effector of kicking leg) to follow the reference curve obtained by the B\'ezier parameters $\alpha$, motion time span $T_\delta$, and current motion phase $t$. 
The rest of the reward terms have a similar formulation as~\eqref{eq:end_effector} and they serve different purposes.
The term $r^m_{c,k}$ and $r^b_{c,k}$ are designed to encourage the other stance legs and the robot base to imitate a nominal standing motion, respectively.
In order to make the robot motion smoother, we include $r^{\dot{m}}_{c,k}$ and $r^{\dot{b}}_{c,k}$ to stimulate the robot to damp out the motor and base velocities. 
$r^{\Delta \mathbf{a}}_{c,k}$ is also introduced to minimize the change of the control policy output $||\mathbf{a}_k - \mathbf{a}_{k-1}||_2$ between two adjacent time step.
Moreover, we add $r^{\tau}_{c,k}$ to minimize the current torque consumption for energy efficiency. 
Additionally, $w_c$ is a normalization weight vector with a dominating weight on the end-effector position tracking term $r^e_{c,k}$.

\begin{table}[t]
\centering
\caption{Randomization range during training.}
\label{tab:randomization}
\begin{tabular}{ccc}
\hline
\textbf{Parameter}            & \textbf{Range}                 & \textbf{Unit}    \\ \hline
\multicolumn{3}{c}{Control and Planning}                 \\ \hline
Robot Link Mass            & $[0.5, 1.5] \times$ default  & kg      \\
Robot Link Inertia         & $[0.7, 1.3] \times$ default  & kgm$^2$  \\
Robot Base Mass Center     & $[-0.1,0.1]$            & m       \\
Robot Link Mass Center     & $[-0.05,0.05]$          & m       \\
Robot Joint Damping        & $[0.7, 4.0]$            & Nms/rad \\
Ground Frictions     & $[0.5, 3.0]$            & 1       \\
Motor Encoder Noise Mean  & $[-0.01, 0.01]$         & rad     \\
Gyro Rotation Noise Mean  & $[-0.01, 0.01]$         & rad     \\
Communication Delay  & $[0, 0.025]$            & sec     \\ \hline
\multicolumn{3}{c}{Only for Control}                            \\ \hline
Standing Time Span $T_{\delta=0}$    & $[1.0, 4.0]$            & sec     \\
Lifting Time Span $T_{\delta=1}$    & $[3.0, 4.0]$            & sec     \\
Kicking Time Span $T_{\delta=2}$    & $[0.2, 0.4]$            & sec     \\
Resting Time Span $T_{\delta=3}$    & $[1.0, 3.0]$            & sec     \\
B\'ezier Parameters $\alpha_{0,1,4}$   & $[-0.1, 0.1]$ + nominal & m       \\
B\'ezier Parameters $\alpha_{2,3}$   & $[-0.1, 0.3]$ + nominal & m       \\
Perturbation Force and Torque  & $[-20,20]$, $[-5,5]$      & N, Nm \\
\hline
\multicolumn{3}{c}{Only for Planning}                           \\ \hline
Ball Stiffness       & $[0.7, 2.0]$            & N/m     \\
Ball Mass      & $[0.5, 1.5] \times$ default   & kg        \\
Ball Inertia and Radius     & $[0.7, 1.3] \times$ default & kgm$^2$, m \\ 
Ball Detection Noise & $[-0.05, 0.05]$         & m       \\
Ball Detection Delay & $[0, 0.3]$              & sec    \\ \hline
\end{tabular}
\end{table}

\subsection{Domain and Motion Randomization}~\label{subsec:control_random}
In order to generalize the control policy in the environments with uncertain dynamics properties, such as in the real world, the dynamics parameters of the robot and the environment are randomized during training in simulation and the range of the uniform randomization is presented in Table~\ref{tab:randomization}. 
For example, in order to encourage the robot to stay robust to the modeling error between the simulation and real world, in each simulation episode, the mass, inertia, and mass center position of each robot link, as well as joint damping ratio, are changed and sampled within recorded range in Table~\ref{tab:randomization}. 
Ground frictions are randomized in order to encourage the robot to be robust to the friction changes in the real world.
Moreover, the sensor noise is simulated as Gaussian distribution with the mean sampled from the presented range, and communication delay between the computer running the RL policy and the low-level computer is also introduced. 
Additionally, we also randomly apply a 6 DoF random perturbation force to the robot base during training in order to increase the robustness of the policy.

Furthermore, as introduced in Sec.~\ref{sec:framework}, we want to develop a control policy that is able to track arbitrary end-effector (toe) trajectories for quadrupedal robots. 
Therefore, as shown in Table~\ref{tab:randomization}, the desired B\'ezier parameters $\alpha$ and time span $T_\delta$ are also randomized during training.
Such a motion randomization is based on nominal hand-crafted lifting, kicking, and resting motions, respectively. 
Based on the B\'ezier parameters of the end-effector trajectory of each nominal motion, we add a large range of randomization of the desired $\alpha$ for the control policy to learn.
Furthermore, the time spans $T_\delta$ for each motion during one soccer shooting maneuver are also randomized. 
In this way, we can encourage the robot to learn a large repertoire of shooting motion.

\subsection{Training Setup}
During training, each episode has a horizon of $N=2500$ timesteps, which is approximately 80 seconds in length. 
In each episode, the robot is repeatedly required to track random shooting motions sampled according to the parameters from Table~\ref{tab:randomization}, and the dynamics parameters are also randomized according to Table~\ref{tab:randomization} and kept fixed over the course of an episode.
After one motion is completed, the robot needs to learn to keep standing until the next new motion begins. 
When the robot falls over, the episode will be terminated and the policy receives 0 return for all remaining timesteps.
In this way, together with the end-effector tracking reward formulated in~\eqref{eq:control_reward} by using the control policy, we can encourage the robot to stay as close as possible to the reference end-effector trajectory while respecting the robot dynamics limitation and motion stability (not falling over). 
The episode will also be terminated if end-effector deviates from the reference by more than a relatively large threshold. 
This can prevent the robot adopting a very conservative behavior, such as just holding the toe in the air. 
The parameters of the control policy is optimized by Proximal Policy Optimization~\cite{schulman2017proximal} to maximize the total expected discounted reward in one episode.

\section{Learning the Soccer Shooting Plan}
\label{sec:planning}
After obtaining a control policy that is able to track arbitrary shooting motions, we now develop a planning policy using RL to decide an optimal motion that can shoot the soccer ball to the goal location in both simulation and the real world.  

\subsection{Training Environment}
\subsubsection{Action Space}
As illustrated in Fig.~\ref{fig:framework}, the action $\mathbf{a}^p_k$ of the planning policy at time step $k$ is the B\'ezier parameters $\alpha \in \mathbb{R}^{3\times5}$ representing the desired robot motion at current time. 
This is sent to the control policy to track.

\subsubsection{State Space}~\label{subsec:planner_obs}
The observation of the planning policy $\mathbf{s}^p_k$ consists of five components.
It includes a goal position $g_k \in \mathbb{R}^2$ which is the $2D$ shooting target on the ground.
The second part is the detected ball position $\hat{o}_k \in \mathbb{R}^3$ relative to the robot base and a history of its positions at last 15 time steps (lasting about half a second).
By including this information, the policy can not only know the current ball position but learn to estimate its velocity and high order information. 
Moreover, the last planner output and past $6$ time step observed robot states $\hat{q}_m \in \mathbb{R}^{12}$, $\hat{q}_{\psi,\theta,\phi} \in \mathbb{R}^3$ are also contained in the observation.
Just like the control policy, we also need to inform the planning policy about the current motion by inputting the motion indicator $\delta \in \mathbb{Z}$.

\subsubsection{Policy Representation}
The planning policy is also represented by a multilayer perceptron which consists of one hidden layer with $256$ units followed by one hidden layer with a size of $128$, and both of them use $tanh$ activation.

\subsection{Reward}
The reward for the planning policy only has one term which is the distance between the ball position $o_k$ and the goal $g_k$, formulated as
\begin{equation}~\label{eq:plan_reward}
    r_{p,k} = 
    \begin{cases}
    1.0, & \text{if goal is reached} \\
    \exp{(-\rho^g\|o_k - g_k\|_2)},& \text{otherwise}
\end{cases},
\end{equation}
where $\rho^g >0$ is a scaling variable. If the ball has reached the goal (within a 0.2-meter range of the goal), the reward $r_{p,k}$ is always $1$. In this way, we can encourage the robot to try its best to shoot to the target. 

\begin{figure}[t]
    \centering
    \includegraphics[width=\linewidth]{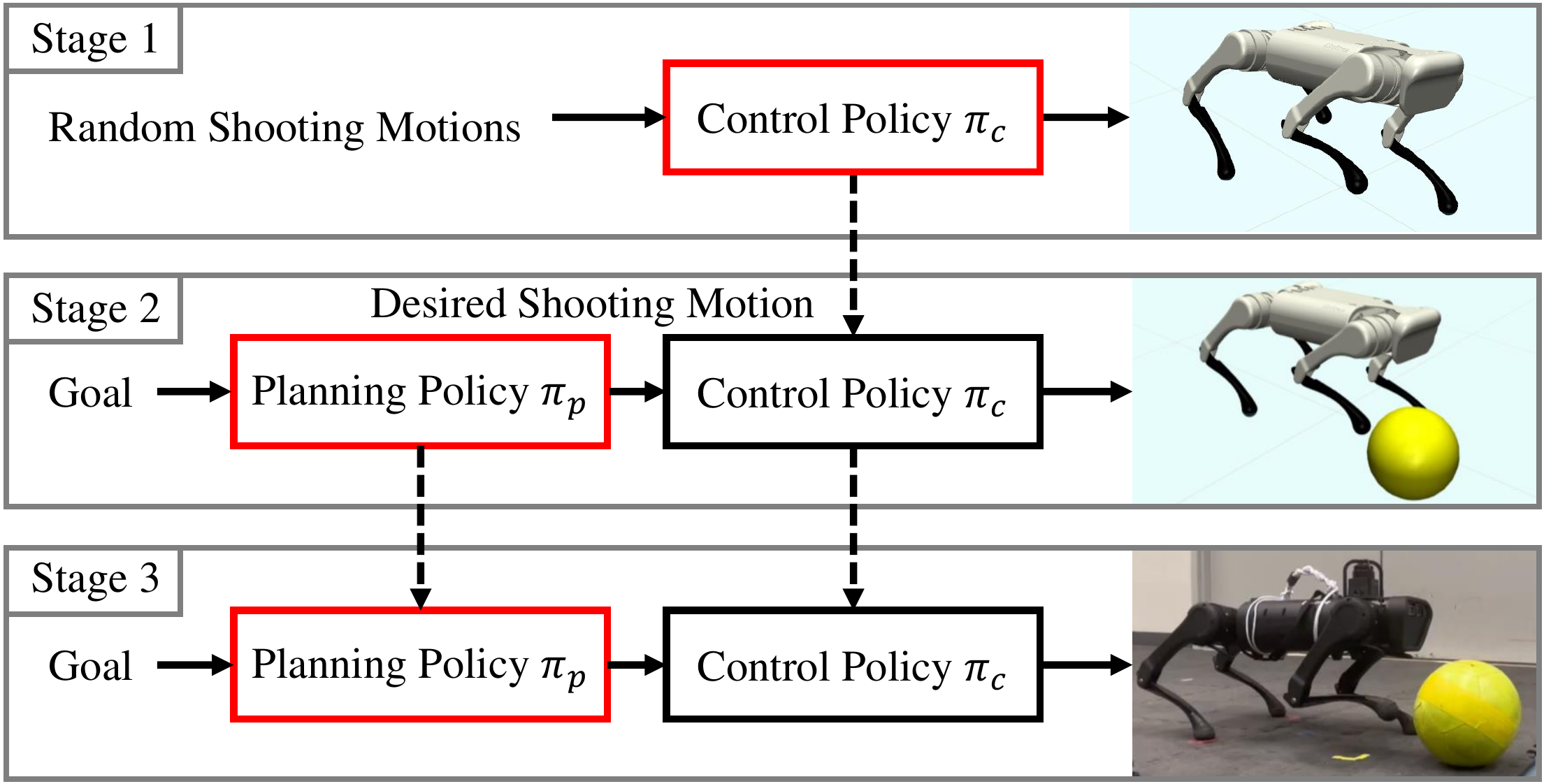}
    \caption{Multiple stage training scheme. Red bounding box represents the policy being optimized at current stage. Stages 1 and 2 are performed in simulation and Stage 3 is in the real world. In Stage 1, a control policy $\pi_c$ is firstly trained to track random shooting motions. It also learns to stay robust to the sim2real transfer by domain randomization. After the control policy is ready, a planning policy $\pi_p$ is pretrained to plan for desired shooting motion for the controller in order to shoot the ball to the goal. Such a planning policy is firstly trained in simulation with a rigid ball in Stage 2 then fine-tuned in the real world to shoot a soft ball in Stage 3.}
    \label{fig:multi_stage}
    \vspace{-0.5cm}
\end{figure}

\subsection{Training Setup}
Real life soccer balls are deformable bodies in the shape of a truncated icosahedron, making it challenging to simulate the contact with the ground and the robot leg.
In order to tackle the gap between the simulation and real world, we adopt a multiple-stage training strategy that includes training in both simulation and the real world, as done in~\cite{smith2021legged}. 
In the training stage for the planner, as presented in Fig.~\ref{fig:multi_stage}, the policy is firstly pretrained with a simulated rigid spherical ball, and then based on the optimal policy parameters obtained in simulation, this policy keeps training in the real world by collecting data from the interaction with the real deformable and icosahedron-shaped ball.

\subsubsection{Pretrain in Simulation}
We utilize MuJoCo to simulate both the A1 robot and a rigid spherical ball to pretrain the planning policy.
During the pretraining stage, each episode lasts around $80$ seconds which is the same as the training stage for the control policy. 
In each episode, the robot is trained to shoot the ball to a random target, and if the ball has not reached the goal after it has stopped or after a given time span, the episode will be terminated to prevent the agent having future return for this shooting failure.
If the ball has reached the goal, the ball position will be reset to a random place next to the robot, and the robot will be required to perform the next round of shooting. 

Similar to training the control policy for the sim2real transfer in Sec.~\ref{subsec:control_random}, dynamics parameters of the training environment for the planning policy is also randomized using the range demonstrated in Table~\ref{tab:randomization} in each episode.   
Besides considering the robot modeling errors and environment changes, we also include the randomness of the simulated ball, such as stiffness, mass, and size.
Moreover, to consider uncertainty of the ball detection algorithm through vision in the real world, we further include a simulated Gaussian noise to the detected ball position and delay into the observation. 

To optimize the parameters of the planning policy, we use Randomized Ensembled Double Q-Learning~(REDQ)~\cite{chen2021randomized} due to its sample efficiency, which is important for fine-tuning in the real world.

\subsubsection{Fine-tuning in Real World}
After the training in simulation has converged, the planning policy is then deployed on the quadrupedal robot to learn the shooting skill in the real world. 
As shown in Fig.~\ref{fig:multi_stage}, the quadruped runs the control policy obtained in Sec.~\ref{sec:control} which is also the same policy used during pretraining in the simulation. 
The planning policy is warm started with the parameters obtained in simulation and keeps training using the same REDQ algorithm but with the samples collected in the real world. 
As shown in Fig.~\ref{fig:main}, the robot hardware is the A1 robot from Unitree Robotics. 
The ball position is detected by a RGB-Depth camera alongside the robot, Intel RealSense D435i camera, by color segmentation, and the goal location is marked by an AprilTag~\cite{apriltag2011} which is also detected by this camera.

The training samples are collected when the robot hardware interacts with the soccer ball in the real life environment. 
The planner's observation introduced in Sec.~\ref{subsec:planner_obs} can be obtained by the robot's onboard sensors and the RGB-Depth camera. 
The reward used in this stage is the same as the one in pretraining~\eqref{eq:plan_reward} to stimulate the agent to shoot the ball to the detected goal location.
During the reset at each episode in the real world, we randomly place the goal location (AprilTag), reset the robot' pose and manually place the ball next to the robot.

\section{Experiments}
\label{sec:result}

\begin{figure}[t]
    \begin{subfigure}{0.99\linewidth}
        \centering
        \includegraphics[width=\linewidth]{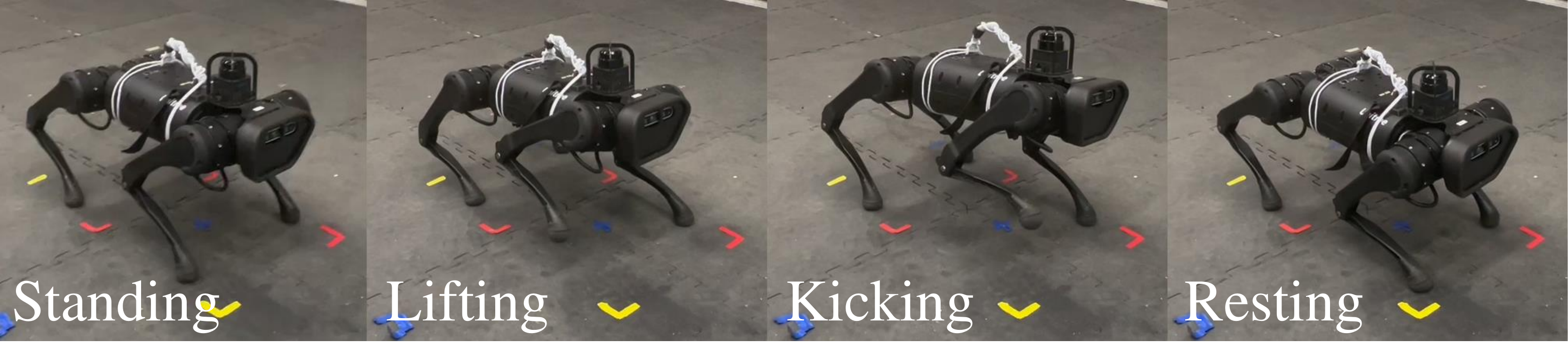}
        \caption{Following a random trajectory} 
        \label{fig:control_nominal}
    \end{subfigure}
    \begin{subfigure}{0.99\linewidth}
        \centering
        \includegraphics[width=\linewidth]{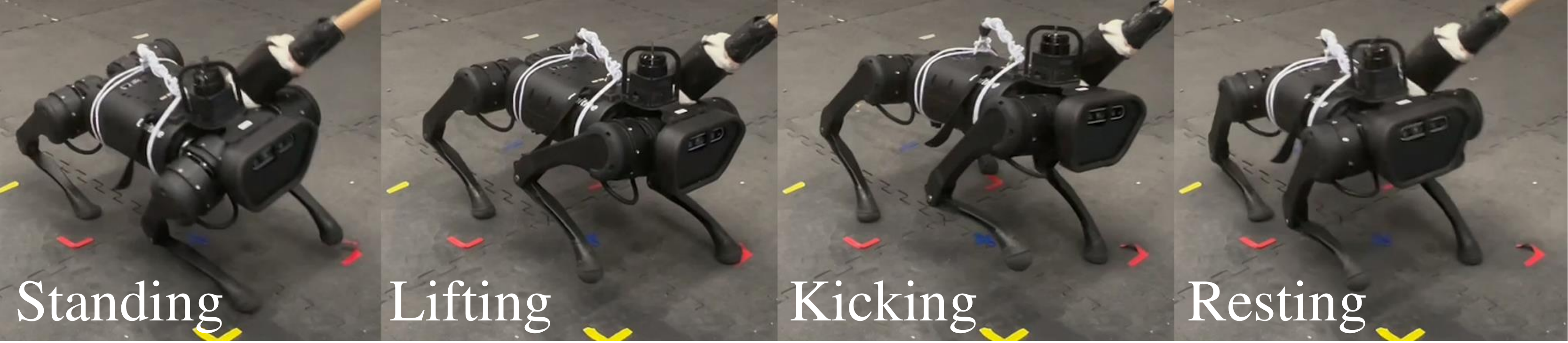}
        \caption{Following a trajectory with random perturbation} 
        \label{fig:control_pertubation}
    \end{subfigure}
    \begin{subfigure}{0.99\linewidth}
        \centering
        \includegraphics[width=0.85\linewidth]{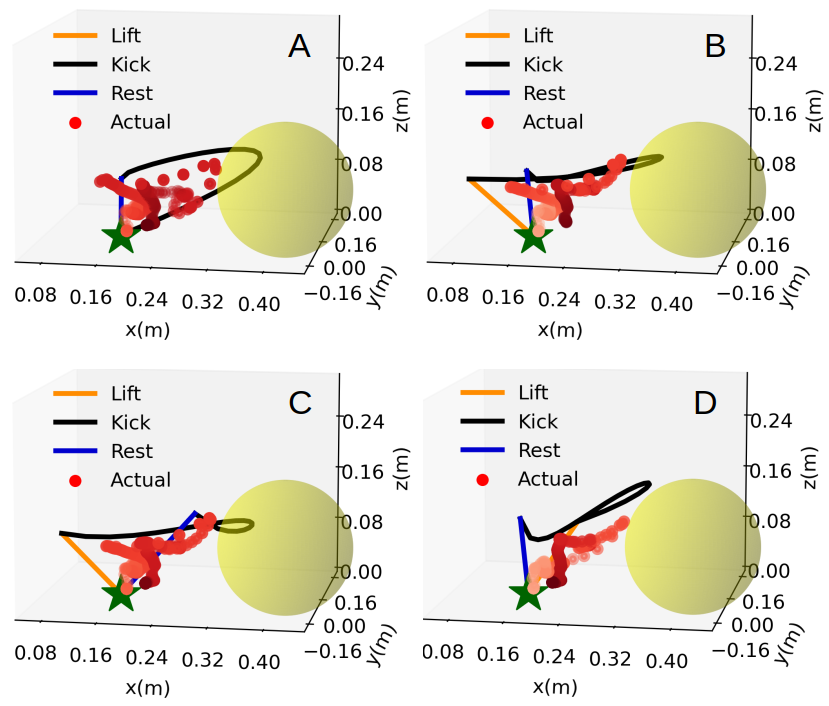}
        \caption{Robot's end-effector (front right toe) tracking a desired trajectory}
        \label{fig:lowlevelperf}
    \end{subfigure}
    \caption{Performance of the control policy deployed on the robot. (a) The control policy enables the robot to track random end-effector (front right toe) trajectories while maintaining balance, even in the scenario (b) where a random perturbation is applied to robot. 
    (c) shows the desired end-effector trajectory (solid lines) obtained from the planning policy in different motions (marked by different colors) in order to shoot the detected ball (marked as yellow) to A) a goal to robot's left, B,C) a goal in front of the robot, and D) a goal to robot's right. The red dashed line depicts actual robot's end-effector position and the darkness of its point represents elapsed time. 
    The green star is the start and the end of the reference trajectory.}\label{fig:control}
    \vspace{-5mm}
\end{figure}

The performance of the control policy on the A1 robot and shooting accuracy of the planning policy before/after fine-tuning in the real world are demonstrated in the accompanying video (\url{https://youtu.be/bteipHcJ8BM}) and analyzed below.

\subsection{Control Performance}
We first validate the performance of the control policy on the real robot. 
During the test, the robot is required to track a random end-effector trajectory using the proposed control policy. 
As shown in Fig.~\ref{fig:control_nominal}, our control policy can be transferred and deployed on the real robot without any tuning.
Furthermore, using the same control policy, by just changing the desired B\'ezier parameters and phase time span $T_\delta$, the robot is able to perform different fast kicking motions while maintaining balance. 
The control policy also shows considerable robustness under random force. 
As demonstrated in Fig.~\ref{fig:control_pertubation}, the control policy is able to prevent the robot falling over by adjusting its stance legs when we perturb the robot randomly. 

Fig.~\ref{fig:lowlevelperf} shows examples of the robot tracking desired end-effector trajectories obtained by the planning policy to shoot a ball to different places.
Please note that the given trajectory (drawn as solid line) is not necessary to be dynamically feasible for the quadrupedal robot. Looking at the robot actual end-effector position marked as red points, the control policy shows the capacity to stay close to the given trajectory while respecting to the dynamics limitation of the robot.

\begin{figure}[t]
    \begin{subfigure}{0.495\linewidth}
        \centering
        \includegraphics[width=\linewidth]{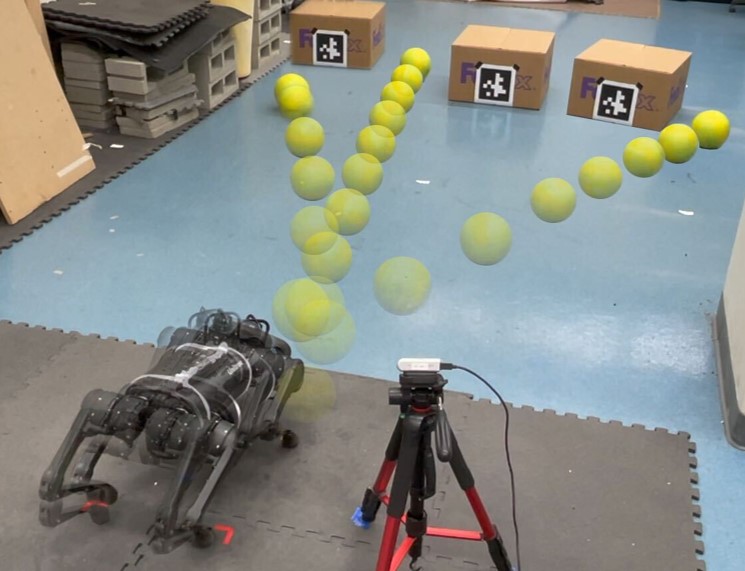}
        \caption{Snapshot before fine-tuning} 
        \label{subfig:snapshot_before_finetune}
    \end{subfigure}
    \begin{subfigure}{0.495\linewidth}
        \centering
        \includegraphics[width=\linewidth]{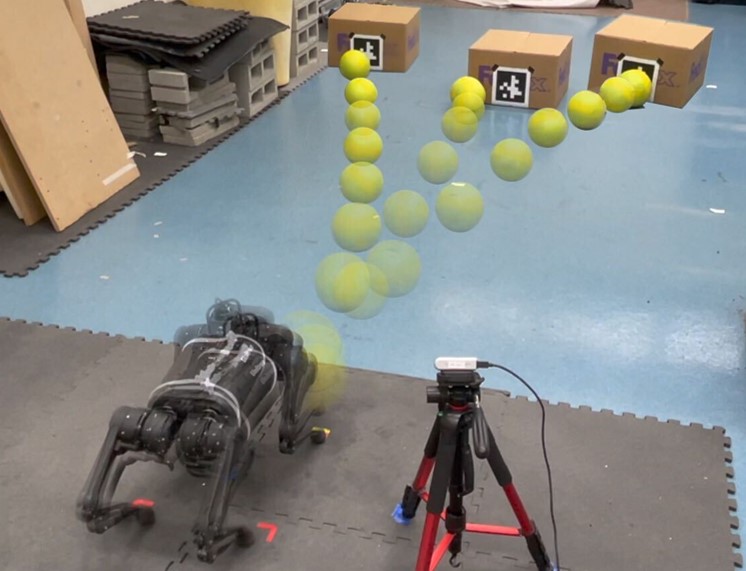}
        \caption{Snapshot after fine-tuning} 
        \label{subfig:snapshot_after_finetune}
    \end{subfigure}    
    \begin{subfigure}{0.495\linewidth}
        \centering
        \includegraphics[width=0.7\linewidth]{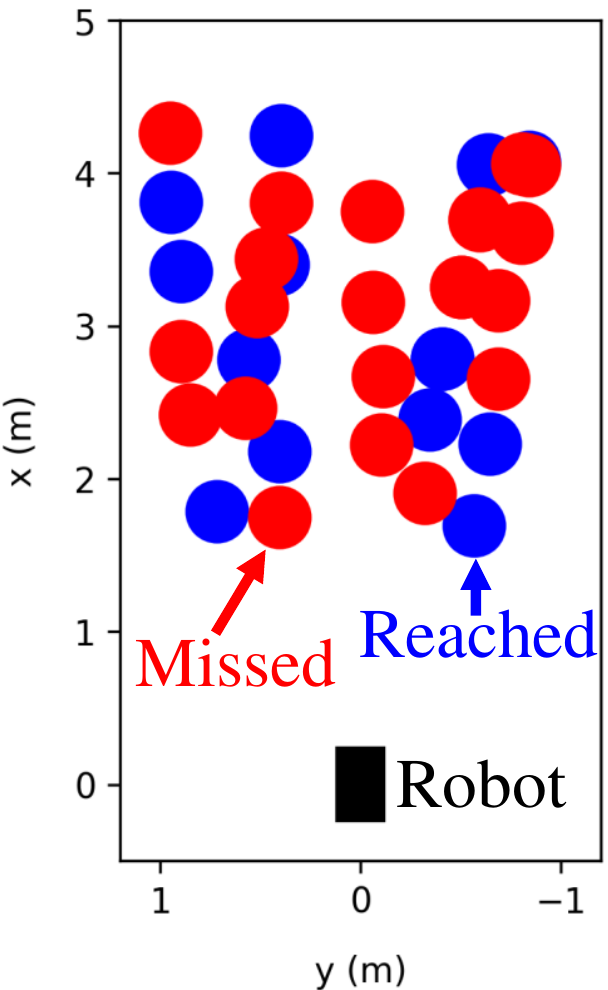}
        \caption{Accuracy map before fine-tuning} 
        \label{subfig:acc_before_finetune}
    \end{subfigure}
    \begin{subfigure}{0.495\linewidth}
        \centering
        \includegraphics[width=0.7\linewidth]{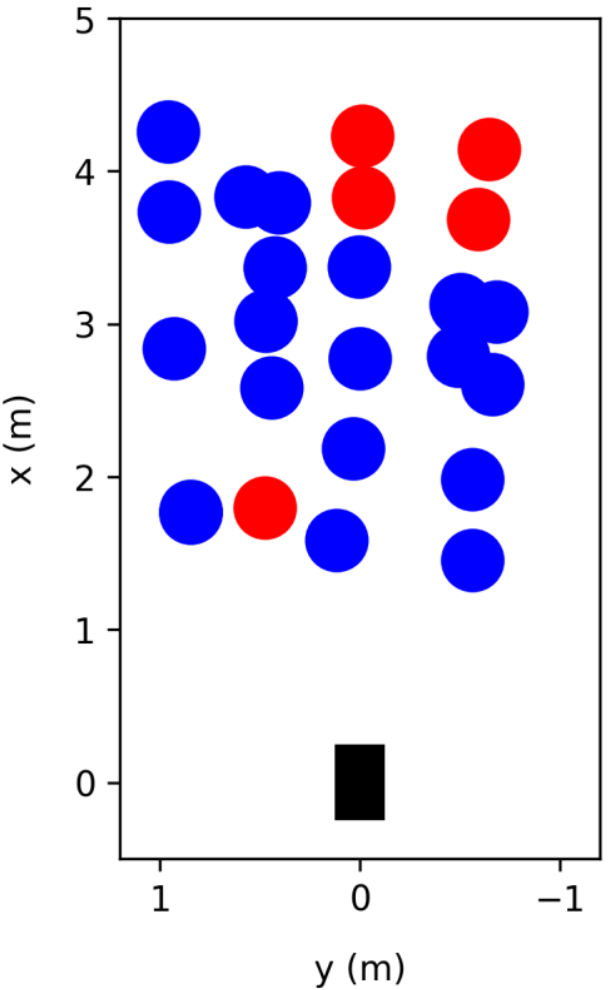}
        \caption{Accuracy map after fine-tuning} 
        \label{subfig:acc_after_finetune}
    \end{subfigure}    
    \caption{Performance of the planning policy before (left) and after (right) fine-tuning in the real world. (a,c) The robot cannot precisely shoot the yellow soccer ball to a random target (the AprilTag on the box) if it uses the planning policy right after pretraining in simulation. Such failure is because of the sim2real gap. (b,d) After we keep training the robot in the real world to kick a real soccer ball, the shooting accuracy improves to a large extent and the robot is able to shoot the soccer ball to most of the region on the map.}
    \label{fig:experiment_snapshot_map}
    \vspace{-0.5cm}
\end{figure}

\subsection{Planning Performance}

After validating the learned control policy on the robot, as illustrated in Fig.~\ref{fig:multi_stage}, we can reuse this well-tested control policy for training the planning policy.
We next conduct experiments with the planning policy, as recorded in Fig.~\ref{fig:experiment_snapshot_map}. 

If we directly use the planning policy pretrained in simulation to shoot the soccer ball to a random target in the real world, the ball can barely hit the target, especially for targets that are over 3 meters away, as demonstrated in Fig.~\ref{subfig:snapshot_before_finetune}. 
This illustrates the huge sim2real gap due to the hard-to-model soft soccer ball, uncertain rolling friction, and contacts with the deformable ball, even though we randomized the simulated ball dynamics parameters in Table~\ref{tab:randomization}.
After training the planning policy with 1386 samples (23-minute data) in 32 iterations in the real world, the precision of the robot shooting skills increase significantly and the robot is able to shoot the ball to the target which can not be reached before fine-tuning, as shown in Fig.~\ref{subfig:snapshot_after_finetune}. 
Please note that for the three locations demonstrated in Figs.~\ref{subfig:snapshot_before_finetune},\ref{subfig:snapshot_after_finetune}, we repeat the shooting experiments to the same locations for three consecutive times, and the ball failed to reach the goal in all of the trials before fine-tuning while the fine-tuned policy enables the soccer ball to reach the goal in 8 out of 9 trials. 

In order to quantitatively analyze the shooting accuracy before and after fine-tuning, we extensively test the planning policy on the robot, and record each trial in Figs.~\ref{subfig:acc_before_finetune},~\ref{subfig:acc_after_finetune}. 
From the recorded data, the shooting accuracy before fine-tuning is $40.6\%$ (13 reach in 32 trials), and is boosted after fine-tuning to $80.8\%$ (21 reach in 26 trials) and the robot demonstrates the capacity to shoot the soccer ball to most of region in the $2\times4.5$ m$^2$ map.  
Such improvement showcases the importance of the fine-tuning stage in Fig.~\ref{fig:multi_stage}. 
However, we note that the resulting planning policy can still not shoot to all the targets, especially for the targets that are far way from the robot as shown in Fig.~\ref{subfig:acc_after_finetune}. 
Such failures may be due to the limitation of maximum torque of the A1 robot.
Interestingly, an emergent behavior shows up in some scenarios, typically when the goal is near to the wall, the robot tends to kick the ball towards the wall and then the ball can bounce off to the goal.

\subsection{Locomotion and Shooting}
We also combine both locomotive skills and shooting maneuver into one task where the robot walks to approach the ball away from it's initial position, switches to standing, and shoots it to a random desired location afterwards.
Experimental results are demonstrated in the video where we use walking controller developed in~\cite{yang2022fast}.
Such experiments showcase the advantages of using quadrupedal robot, which can uses its leg to not only walk but to manipulate the ball, over the robotic arm which has a fixed base.

\section{Conclusion and future works}
In conclusion, we demonstrate a hierarchical reinforcement learning framework to enable precise soccer shooting skills on quadrupedal robots.
In this work, we decompose the soccer shooting problem into two sub-problems: motion control to perform arbitrary shooting motions using one leg while balancing on the others and motion planning to find an optimal motion to shoot the soccer ball to the target.
By separating the high-level soccer shooting problem, we are able to firstly focus on developing a robust control policy that enables the robot to track arbitrary shooting motions in the real world and use it for training the motion planning policy.
As a real soccer ball is deformable and its contact is hard to simulate, we leverage multi-stage learning to firstly train the planning policy in simulation with a rigid ball and keep training the policy on the real robot shooting a real soccer ball. 
In experiments, we demonstrate the capacity of the control policy to enable a quadrupedal robot to track a random but fast shooting motion while staying robust to sim2real transfer and random perturbations. 
After fine-tuning the planning policy in the real world, the robot is able to reliably shoot the soccer ball to random targets with the proposed framework. 
Note that this work only focuses on the soccer shooting maneuver when the quadrupedal robot is standing. In the future, it will be interesting to extend such a method to combine quadrupedal locomotion and ball manipulation skills to perform more complex soccer skills.

\section*{Acknowledgements} 
This work was supported in part by Hong Kong Centre for Logistics Robotics and in part by NSF Grant CMMI-1944722. The authors would also like to thank Laura Smith and Hongbo Zhang for their gracious help.

{
\balance
\bibliographystyle{IEEEtran}
\bibliography{bib/bibliography}
}

\begin{acronym}
\acro{HP}{high-pass}
\acro{LP}{low-pass}
\end{acronym}

\end{document}